\documentclass[letterpaper, 10 pt, conference]{ieeeconf}  % Comment this line out if you need a4paper
\let\labelindent\relax
\usepackage[font=small,labelfont=bf]{caption}
\usepackage{subcaption}
\usepackage{wrapfig}
\usepackage{microtype}
\usepackage{xspace}
\usepackage{epsfig}
\usepackage{graphicx}
\usepackage{amsmath, bm}
\usepackage{amssymb}
\usepackage{algorithm}
\usepackage{algorithmic}
\usepackage{mathtools}
\usepackage{array}
\usepackage{multirow}
\usepackage[inline]{enumitem}
\usepackage{multirow}
\usepackage{tabularx}
\usepackage{color}
\usepackage[utf8]{inputenc} % allow utf-8 input
\usepackage[T1]{fontenc}    % use 8-bit T1 fonts
\usepackage[hyphens]{url}   % simple URL typesetting
\usepackage{booktabs}       % professional-quality tables
\usepackage{amsfonts}       % blackboard math symbols
\usepackage{nicefrac}       % compact symbols for 1/2, etc.
\usepackage{xcolor}
\usepackage{rotating}
\usepackage{lipsum}
\usepackage{wrapfig}
\usepackage{flushend}
\usepackage{cool}
\usepackage{siunitx}
\usepackage{bbm}

\usepackage[capitalise,nameinlink]{cleveref}

\makeatletter
\DeclareRobustCommand\onedot{\futurelet\@let@token\@onedot}
\def\@onedot{\ifx\@let@token.\else.\null\fi\xspace}
\makeatother

\def\etc{\emph{etc}\onedot}
\def\ie{\emph{i.e}\onedot}
\def\eg{\emph{e.g}\onedot}
\def\cf{\emph{cf}\onedot}
\def\vs{\emph{vs}\onedot}

\def\R{\mathbb{R}}

\definecolor{redcol}{rgb}{1, 0, 0}
\definecolor{bluecol}{rgb}{0, 0, 1}

\renewcommand{\paragraph}[1]{\smallskip\noindent{\bf{#1}}}
\newcommand{\mast}[0]{MaAST\xspace}
\newcommand{\spl}{\texttt{SPL}\xspace}

\newcommand{\rgb}{\texttt{RGB}\xspace}
\newcommand{\depth}{\texttt{Depth}\xspace}
\newcommand{\rgbd}{\texttt{RGB-D}\xspace}
\newcommand{\rgbdocc}{\texttt{RGBD+OCC}\xspace}
\newcommand{\rgbdexp}{\texttt{RGBD+EXP}\xspace}
\newcommand{\rgbdsem}{\texttt{RGBD+SEM}\xspace}

\newcommand{\occatt}{\texttt{RGBD+OCC+ATT}\xspace}

\newcommand{\stopaction}{\texttt{STOP}\xspace}
\DeclareSIUnit{\million}{\text{milion}}
\def\@fnsymbol#1{\ensuremath{\ifcase#1\or *\or \dagger\or \ddagger\or
   \mathsection\or \mathparagraph\or \|\or **\or \dagger\dagger
   \or \ddagger\ddagger \else\@ctrerr\fi}}
\IEEEoverridecommandlockouts                              % This command is only needed if 
\usepackage[noadjust]{cite}
\title{\LARGE \bf
\mast: Map Attention with Semantic Transformers \\ for Efficient Visual Navigation
}

\author{Zachary Seymour$^{\dag}$ Kowshik Thopalli$^{*}$ Niluthpol Mithun$^{\dag}$ Han-Pang Chiu$^{\dag}$
Supun Samarasekera$^{\dag}$ Rakesh Kumar$^{\dag}$% <-this % stops a space
\thanks{$^{\dag}$SRI International; Email: 
        {\tt\scriptsize firstname.lastname@sri.com}}%
\thanks{$^{*}$Arizona State University; Email: 
        {\tt \scriptsize kthopall@asu.edu}}%
}

\begin{document}

\maketitle
\thispagestyle{empty}
\pagestyle{empty}

\begin{abstract}
Visual navigation for autonomous agents is a core task in the fields of computer vision and robotics. Learning-based methods, such as deep reinforcement learning, have the potential to outperform the classical solutions developed for this task; however, they come at a significantly increased computational load. Through this work, we design a novel approach that focuses on performing better or comparable to the existing learning-based solutions but under a clear time/computational budget. To this end, we propose a method to encode vital scene semantics such as traversable paths, unexplored areas, and observed scene objects--alongside raw visual streams such as RGB, depth, and semantic segmentation masks---into a semantically informed, top-down egocentric map representation. Further, to enable the effective use of this information, we introduce a novel 2-D map attention mechanism, based on the successful multi-layer Transformer networks. We conduct experiments on 3-D reconstructed indoor PointGoal visual navigation and demonstrate the effectiveness of our approach. We show that by using our novel attention schema and auxiliary rewards to better utilize scene semantics, we outperform multiple baselines trained with only raw inputs or implicit semantic information while operating with an $80\%$ decrease in the agent's experience. 
% \red{Thus, our approach directly translates into providing significant performance gains while decreasing training time, computational requirements, and cost. 
% Our experiments show that with these changes, agents trained with our method perform much better as compared to agents trained with only raw inputs or implicit semantic information.}
\end{abstract}
\section{Introduction}

Humans employ semantic scene structures to reason about the world and pay particular attention to relevant semantic landmarks to develop navigation strategies~\cite{braincognition, humannavigation}.
Therefore, humans can efficiently learn from past navigation experience to explore new environments, such as discovering common semantic scene entities that can be generalized to similar but previously unseen places.   
\begin{figure}
\vspace{0.1cm}
    \centering
    \includegraphics[width=0.45\textwidth, height=0.3\textwidth]{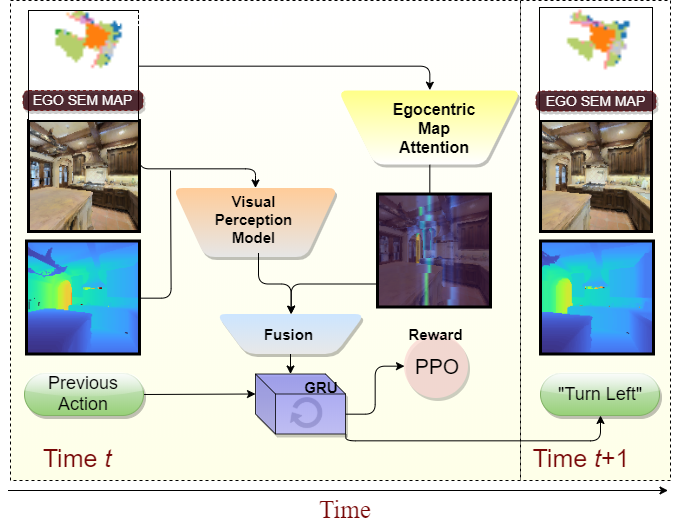}
    \vspace{-0.2cm}
    \caption{We show an overall diagram of our proposed approach, \mast. Given \rgb and \depth inputs, as well as an accumulated top-down egocentric semantic map constructed by our system, the model uses features from both the visual perception model and our novel map attention module to predict its next action. In addition to the input images, we also visualize the learned semantic map attention, projected back onto the \rgb observation.}
    \label{fig:arch}
    \vspace{-0.2cm}
\end{figure}
The ultimate goal of visual navigation for autonomous agents is to enable agents to resemble or even exceed these human capabilities.
A traditional approach to this problem is to localize the agent in a map of the environment that is built beforehand or constructed using simultaneous localization and mapping (SLAM) algorithms~\cite{thrun2005probabilistic, davison1998mobile, desouza2002vision}. 
Paths are then planned to convey the agent to target locations~\cite{lavalle2000rapidly,canny1988complexity,kavraki1996probabilistic}. 
These geometry-based methods require extensive hand-engineering, and cannot be easily applied to new unexplored environments. 

Recently, learning-based methods---with deep reinforcement learning (DRL) leading the way, are being developed to address this problem.
DRL methods directly learn a mapping from observations to actions through trial-and-error interactions with its environment. 
They have demonstrated promising results on navigation tasks and showed superior performance compared to geometry-based methods~\cite{habitat19iccv}. 
However, current state-of-the-art DRL methods require training with massive amounts of observations to achieve satisfactory performance. They lack the semantic reasoning capabilities that humans possess to quickly learn from past experience and extrapolate to new environments. 
Hundreds of millions of interactions with the training environments, requiring enormous computational resources and distributed processing~\cite{ddppo}, are mandatory to converge to a reasonable policy for complicated scenarios. 
%\red{Millions of interactions with the training environments, requiring enormous computational resources  are mandatory to converge to a reasonable policy for complicated scenarios ~\cite{habitat19iccv}. }

% \red{Another point to add even though the above implicitly implies is: current RL agents start with random actions and hopefully converge to better $\pi$ and utilize no priors at all. Our method is in fact a model that explicitly uses priors and hence better performance in shorter time periods.}

In this paper, we take a step towards more efficient learning, as humans do: learning to achieve comparable (or better) visual navigation performance for autonomous agents with less available budget. 
The budget can be in terms of the number of data observations available to the agent, or the available computational resources (and, thus, time) available for training the model. 

To achieve this objective, we propose to synergistically utilize the scene semantics along with a novel structured attention mechanism in a DRL model, which we call \textbf{Ma}p \textbf{A}ttention with \textbf{S}emantic \textbf{T}ransformers, or \mast (\Cref{fig:arch}).
We propose to first construct a memory structure
%---representing a top-down, egocentric view of the environment that is progressively updated as the agent explores the area---
in which we encode the semantic class of each object (or region) the agent perceives in the environment.
We also introduce a novel formulation of the popular Transformer architecture~\cite{vaswani17:attention}, which we refer to as egocentric map Transformer, to encourage the agent to learn pertinent objects and regions based on their spatial relationship with the agent.
This attention module automatically focuses on relevant parts of the map memory structure so as to determine the next action to take.
%taken by the agent.

Through this work, we make the following observations:
\begin{enumerate*}[label=\roman*)]
    \item a na\"ive data-driven approach of utilizing scene semantics provides no added benefit to using an egocentric occupancy map~\cite{exp4nav};
    \item on the other hand, we show that with our novel attention mechanism, we outperform multiple state-of-the-art approaches particularly in terms of path efficiency; and
    \item through systematic ablation experiments, we establish empirically the superior efficacy of our proposed attention mechanism on semantic features. Our observations corroborate our hypothesis: just as humans do, autonomous agents should pay careful attention to semantic landmarks to obtain improved performance. 
\end{enumerate*}
Thus, the contributions of our proposed approach, \mast, are as follows:
\begin{enumerate}
    %\setlength{\textfloatsep}{2pt plus 2.0pt minus 1pt}
    %\item To the best of our knowledge, we present the first study of using both top-down area maps and coverage rewards for training an reinforcement learning policy end-to-end for the task of visual navigation in unseen environments.
    \item To the best of our knowledge, we present the first study on the effective use of perceived scene semantics and their spatial arrangement to train an agent end-to-end on the task of visual navigation in unseen environments.
    \item We introduce a novel 2-D map attention mechanism, based on Transformers, for encouraging model to extract features from and to focus on the most relevant areas of structured, egocentric, top-down spatial information.
    \item We demonstrate significant improvements in the policy's performance metrics and sample efficiency when learned under restricted training budgets.
\end{enumerate}

\section{Related Work}
%We briefly review prior works related to different perspectives from our proposed method to visual navigation for autonomous agents, with the focus on methods using DRL. 
% We will also review works that exploit attention for improved performance in computer vision and natural language processing tasks.

\textbf{DRL for Navigation:} 
There is a recent surge of interest in using learning-based methods and DRL for visual navigation, and consequently, a number of methods have been proposed~\cite{jaderberg2016reinforcement, pathak2017curiosity,gupta2017unifying,mnih2016asynchronous,exp4nav,AI2THOR,dosovitskiy2016learning,ddppo,anm20iclr}. 
Due to the prohibitive costs associated with physically training agents with DRL, most of the methods instead train agents in a simulator with multi-threaded implementations. 
To this end, several simulators, predominantly supporting indoor navigation, have been proposed~\cite{habitat19iccv,MINOS,brockman2016openai,Gibsonenv,AI2THOR, vizdoom,chalet}.
%, while, \eg CARLA~\cite{CARLA}, an open-source simulator for urban driving systems is proposed to aid in the research of autonomous driving. 
Similarly, several 3D datasets have been collected for indoor navigation. 
Many are synthetic~\cite{SUNCG,stanfordscenes,scenenet}, based on CAD models and hence lacking realism, while others~\cite{Matterport3D, Gibsonenv,stanford2d3d} consist of 3D scans of real environments. 
Thus, many recent works in visual navigation (including our own) focus on these photo-realistic datasets~\cite{habitat19iccv,Mishkin2019BenchmarkingCA,MidLevelVisualRep,situationalfusion}.

\textbf{Exploration and Mapping in DRL:} 
It has been shown that a key factor for success in navigation is exploration; \ie, to increase an autonomous agent's coverage of a given area with efficacy~\cite{yamauchi1997frontier,thrun1999minerva}.
While \cite{yamauchi1997frontier} uses heuristics and \cite{thrun1999minerva} depends on a  human controller to build maps and explore, recent works~\cite{exp4nav,savinov2018episodic,gupta17:cogmapping,anm20iclr} propose solutions to this task of exploration via learning.
Our work is most closely related to recent works~\cite{anm20iclr,exp4nav,occant,chaplot20objectgoal} that address the problem of task-agnostic exploration, using DRL to learn effective exploration policies that maximize the coverage of a given environment.
However, a key drawback in the use of top-down occupancy map~\cite{anm20iclr,exp4nav} is a failure to encode the diverse semantic cues such as doors, different rooms, \etc that humans use to explore or navigate.
While several recent works have begun to show promise in leveraging a hierarchical DRL policy combined with an analytical planner, our focus remains on improving the efficiency of end-to-end DRL solutions by improving the intermediate scene representation.
% Their method first constructs top-down occupancy maps by projecting the observed scene depth, using known camera parameters. 
% An egocentric transformation of this map, along with the agent's current RGB observation, is then used as input to a visual perception model to learn exploration policies.
% However, a key drawback in the utilization of this top-down occupancy map is that it fails to encode the vast diverse semantic cues such as doors, different rooms, \etc that humans use to explore or navigate.
We propose a novel solution which uses segmentation masks of the current observation to accumulate top-down, egocentric semantic maps.
However, we find the richness contained in these maps adds additional complexity for the policy network to properly make use of these features.
Consequently, we also introduce a novel multi-layer Transformer architecture for processing such 2-D top-down structured information, allowing more relevant features to be extracted.

\textbf{Semantic Feature Fusion:}
It has been shown in prior works that semantic cues leads to improved performance in learning-based navigation methods \cite{Yang2018VisualSN,MidLevelVisualRep} as would intuition confirm it to be. However, these methods use the semantically-segmented images as direct inputs to the model, which directly discards information about the 3D structure of the scene.
By contrast, we propose to use those cues to build better occupancy maps and thus a better memory structure for our policy, as opposed to extending RL agents with specialized memory structures such as \cite{NeuralMap,Oh2016ControlOM}.

% There have also been several recent works in the DRL literature incorporating attention for visual navigation tasks, such as using a Transformer over the history of observations to accommodate long-horizon tasks~\cite{SceneMemoryNetworks} or using pairwise semantic task affinities to regularize the decision making~\cite{situationalfusion}.
% In its use of spatial attention over observations of the environment, our work is most similar to recent work in embodied vision-and-language navigation (VLN),
% % In the VLN task, as opposed to ours, 
% in which the agent must navigate to its goal given a set of natural language instructions.
% In~\cite{EmbodiedVLN}, the spatial attention is computed using dynamic convolutional features constructed from attention over the natural language instructions, given the previous action.
% That is to say, there is a known \textit{a priori} correspondence between the vision and language modalities from which the attention is learned. 
% In our case, there exists no such correspondence, and thus egocentric self-attention is learned to focus on the most relevant areas of the map without such supervision.

\section{Approach}
%In this section, we describe our proposed approach to improve utilization of scene structure and semantics for performing learning-based visual navigation.

%We begin by outlining the problem definition, along with our construction of top-down, egocentric semantic maps. We then layout our baseline policy architecture and introduce our novel multi-layer Transformer formulation, that extracts relevant features from the aforementioned structured map data.

\paragraph{Problem Setup: }In this work, we consider the task of PointGoal navigation; \ie, given a specified starting location and orientation, the agent is required to navigate to the desired target location in an unseen cluttered environment. However, our proposed method is general and can be easily extended to other navigation tasks, such as ObjectGoal and AreaGoal tasks. 
We refer our readers to \cite{Anderson2018OnEO} for a concise description of these tasks.
%For the purposes of our experiments, 
We assume that our navigation agents are equipped with an \rgb camera and a \depth sensor. We also assume the agent has a semantic sensor, capable of producing semantic segmentation of the current \rgb camera observation - a class number $c \in \{1,\dotsc, N_c\}$ for each pixel in the RGB image. Note our goal is to show increased learning efficacy of the agent, given an improved understanding of the scene. 

Our objective is to learn a policy $\pi$ through reinforcement learning that makes effective use of all three sensory outputs (\rgb camera, \depth sensor, and the semantic sensor) for navigation.
To achieve this, we propose a novel method to first construct and maintain an egocentric semantic map of the environment using the \depth and semantic sensors. We construct an image of the map such that the agent is at the center, and the map is rotated such that the agent's heading points upwards.
Previous work has shown that providing autonomous agents a local map of its surroundings enables stronger reward signals to reinforce learning and encourages exploration of new environments~\cite{exp4nav}.
Our intuition is that enhancing the map with a structured semantic representation of the objects and regions in a scene will allow the agent to make better decisions with regards to its goal, in addition to exploring unseen areas.
We finally propose a novel structured attention mechanism to extract information from the map, such that the policy network learns to focus on most relevant regions in the map to predict its next action.

%\vspace{-0.1cm}
\subsection{Egocentric Semantic Map Construction}
\label{sec:egosem}
%\vspace{-0.1cm}

For map construction, we primarily follow~\cite{exp4nav}, in the setting where we assume no estimation error on the agent's pose.
Concisely, when an agent takes an action $a_t$ at time step $t$, it receives a set of observations $O_{t+1} = (r_{t+1}, d_{t+1}, s_{t+1})$ representing the outputs from the \rgb, \depth, and semantic sensors, respectively.
The pixels in the depth image can be back-projected to a 3-D point cloud using the known camera intrinsic parameters.
The extrinsic parameters can then be calculated from the agent's current pose estimate to transform the point cloud from the camera's reference frame to the world frame.
The points are then projected down to form a 2-D map of the environment.
Finally, the map is cropped to a fixed radius $r$ around the agent such that the agent is at the center, of the image and is rotated such that the agent's heading points upwards.
In~\cite{exp4nav}, three types of information are stored in the map.
Any points above and below some height threshold are projected down and stored in the map as obstacles.
The remaining points are stored as traversable space, and the rest of the map is treated as yet unexplored.

Information about traversable and non-traversable space is extremely useful to support path planning in autonomous agents; however, our intuition is that a large amount of this information, particularly for short-term action decisions, is encoded in depth image already, as evidenced by the strong performance of agents with only a depth sensor~\cite{Mishkin2019BenchmarkingCA}.
Although exploration and map-building allow the learned policy to better support back-tracking and movement to unexplored areas,
% the space between walls indicating an egress point is encoded in the same way as a gap between furniture, whether in the occupancy grid or in the local depth image.
we propose that discriminating different objects and free space is more useful for long-term decision making and reward prediction, particularly in large environments where an agent must traverse several rooms to achieve its goal.

\begin{figure}
\vspace{0.15cm}
    \centering
    \includegraphics[width=0.47\textwidth, height=0.25\textwidth]{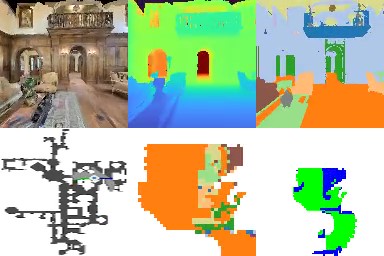}
    \vspace{-0.07cm}
    \caption{An example of the observations available to the agents. Clockwise from top left: \rgb; \depth, colorized with Turbo~\cite{turbo}; semantic segmentation mask; occupancy map, where green denotes traversable space and blue obstacles; semantic map; a ground truth map of the traversed path (for visualization purposes only).}
    \vspace{-0.1cm}
    \label{fig:observations}
\end{figure}

Our approach is to enhance the information stored in the agent's map of the environment with the semantic classes of perceived objects and components in the scene.
That is, for each point categorized as an obstacle above, we instead store its semantic class (\eg, the highest prediction class conditional probability of the pixel from semantic segmentation or value given from the semantic sensor) directly in the map.
When there are instances of vertically overlapping objects (\eg, lamp hanging over a table) or if the map resolution is low, some points may be projected down into the same map location, resulting in a loss of information where one class overwrites others. As such, we instead treat each location in an $2r\times2r$ map $M$ as a binary vector $M_{ij} \in \{0,1\}^N_c$ where, $k = 1$ if class $k$ is present at that location and $0$ otherwise. In this way, as the agent moves through the environment, it accumulates a map containing a bag-of-classes at each grid location. In addition to providing a richer source of information about the environment, this semantically enhanced map allows the same policy architecture to transit easily to other navigation tasks (e.g., ObjectGoal) where the agent must identify semantics of its surroundings for successful goal achievement.
\Cref{fig:observations} shows an example, providing a visualization of the various observation modalities available to the agents.

\vspace{-0.1cm}
\subsection{Policy Architecture}
%\vspace{-0.1cm}

\newcommand{\fcnn}{f_{\text{CNN}}}
\newcommand{\fmap}{f_{\text{MAP}}}
An overall diagram of our policy architecture is shown in \Cref{fig:arch}.
The network receives two sources of information, an \rgbd observation and a 2-D egocentric semantic map. Each of these is processed by a convolutional neural network (CNN) to produce visual and map feature vectors, respectively.
We denote the \rgb and \depth observations at time step $t$ by $(r_{t}, d_{t})$. 
The visual model, denoted $f_{CNN}$, transforms  $(r_{t}, d_{t})$ to $(r^{emb}_{t}, d^{emb}_{t})$ by $\fcnn(r_{t}), \fcnn(d_{t})$.
In our experiments, we concatenate both \rgb and \depth observations and transform them by $\fcnn$, obtaining the corresponding output as ${rd}^{emb}_{t}$; for brevity and without loss of generality, we use $r^{emb}_t$ to denote the same.
While in prior work---where a policy is learned exclusively for exploration~\cite{exp4nav}---each $\fcnn$ was implemented by a ResNet18~\cite{Resnet} CNN pretrained on Imagenet, we found the simpler, 3-layer CNN described in \cite{habitat19iccv} to give better performance for navigation-oriented tasks.

We represent by $\fmap$ the network that transforms the egocentric semantic map at time step $t$, $M_{t}$ (\Cref{sec:egosem}).
We consider two formulations of $\fmap$. In the first case, $\fmap$ uses the same architecture as $\fcnn$: a 3-layer CNN.
This mirrors the method used in~\cite{exp4nav}, where the visual and map information are processed by two versions of the same network with different weights.
However, because the information stored in our semantic map is much richer than that in the occupancy map, we find that the same CNN architecture is unable to adequately extract meaningful representations from it.
Drawing inspiration from prior work in neural mapping and memory~\cite{NeuralMap}, we instead design a network for processing the map input by treating it like a memory unit.
Unlike in~\cite{NeuralMap}, we would like to explicitly preserve the geometry and layout of the scene. We do not need to learn a specialized method to ``write'' to the memory. We only need to ``read'' (or extract feature) from it.
To this end, we utilize a form of Transformer architecture ~\cite{vaswani17:attention, BERT}, adapted to use both a two-dimensional self-attention mechanism~\cite{selfatt} and to learn specialized positional embeddings to better take advantage of the structured nature of the map input.
It allows the network to better learn an representation of semantic classes present in the scene and to extract information about the relevant regions of the map.  
We next present our map attention mechanism, which is designed for this problem.

\vspace{-0.1cm}
\subsection{Egocentric Map Transformer}
\vspace{-0.05cm}

\label{sec:attention}
\label{sec:policy}
Our map attention mechanism is chiefly inspired by the use of the Transformers in natural language processing, which are deep neural networks consisting of multiple layers of multi-headed self-attention.
In the case of natural language (\cf~\cite{vaswani17:attention, BERT}), the self-attention is generally computed using feature representations of each subword in a sentence, computed as the sum of two learned embeddings: a token embedding, which is specific to each subword, and a positional embedding, indicating the position of the token in a sentence. This composition allows the subword representation to vary depending on its location in the sentence.

\begin{wrapfigure}{R}{0.23\textwidth}
    \centering
    \includegraphics[width=0.23\textwidth]{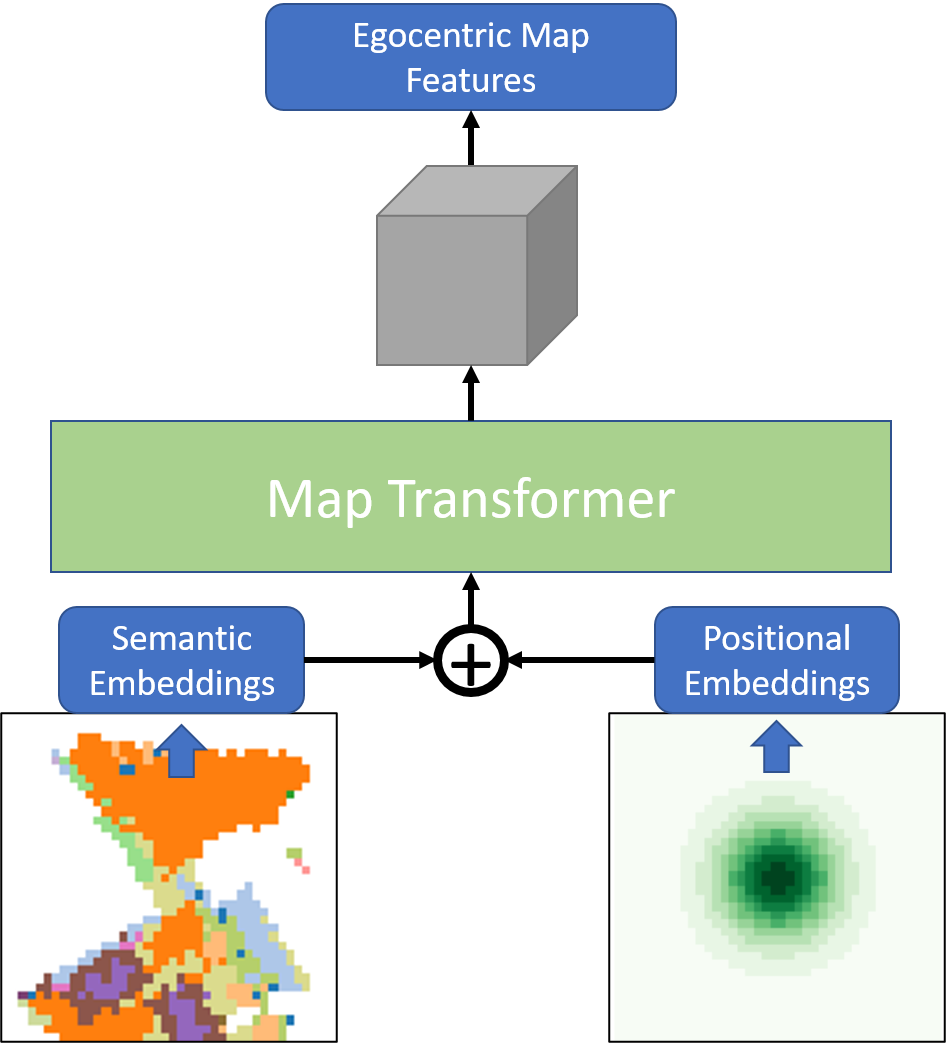}
    \vspace{-0.1cm}
    \caption{An overview of proposed map Transformer module. The input semantic map (color-coded by class) is on the left and, the encoded positional indices (with colors becoming darker closer to the agent) is on the right.}
    \label{fig:attn}
    \vspace{-0.1cm}
\end{wrapfigure}

Prior works using spatial or image-based self-attention have primarily operated on images~\cite{selfatt} (where there is no direct analogue for the positional embedding) or on a sequence of image region features (in which case position encodings are drawn from alignment with corresponding words~\cite{visualbert} or bounding box ~\cite{vilbert}).
However, because the input to $\fmap$ consists of an $2r \times 2r \times N_c$ map centered on the agent, each map cell has important structural information. It is important to encode information about a cell based on its current relationship to the agent.
Namely, for a map of radius $r$, we propose to construct a $2r \times 2r$ matrix of position indices $\mathcal{P}$ using by a Gaussian kernel, centered on the agent, with a scale factor of $\frac{r}{2}$.
As more of the mass of the Gaussian kernel is concentrated around the center, this has the benefit of giving fine-grained positional representations to map cells nearer the agent than those farther away (\Cref{fig:attn}).

Given such a kernel matrix, we construct $\mathcal{P}$ by assigning each unique value in the kernel a unique consecutive integer, which are then used to index the learned positional embeddings.
Thus, the sum of these embeddings and the bag-of-embeddings for the semantic classes at each cell are used as the input to the multi-layer Transformer. 

We adapt each layer of the Transformer to utilize $1 \times 1$ convolutions to preserve the spatial dimensions of the map before pooling across both spatial dimensions and the final output. We denote this final transformed feature representation $\fmap(M_{t})$ by $M^{emb}_{t}$, with both $r^{emb}_{t}, M^{emb}_{t}\in \R^{N_{h}}$ and $N_h$ is the hidden dimension of the policy network. The pooled outputs of $\fcnn$ and $\fmap$ are finally concatenated, and fused by a fully-connected layer. The output along with the agent's previous action is fed into a one-layer gated recurrent unit (GRU)~\cite{cho14:gru}, that enables the agent to maintain coherent action sequences across several time steps (the actor) followed by a linear layer (the critic) to predict the next action.

%\vspace{-0.05cm}
\subsection{Rewards and Optimization}
\label{sec:rewards}
%\vspace{-0.05cm}

The network is trained end-to-end using reinforcement learning, particularly proximal policy optimization (PPO)~\cite{ppo}, following~\cite{habitat19iccv}.
All policies trained in this manner receive a reward $R_t$ at time step $t$ for moving closer to the goal
% \begin{equation}
%         R_t=\begin{cases}
% s + \Delta_{t - 1} - \Delta_{t} + \lambda & \text{if goal is reached}
% \\ \Delta_{t - 1} - \Delta_{t} + \lambda & \text{otherwise}
% \end{cases},
% \end{equation}
% where $s$ is the reward for a success, $\Delta_t$ is the agent's distance from the goal at time step $t$, and $\lambda$ a small time penalty.
equal to the change in distance to the goal plus a large reward $s$ if the goal is reached and a small time penalty $\lambda$.
For agents with access to a map of the environment, we also add an auxiliary reward equal to the percentage of map tiles that have been revealed, to encourage both exploration and better use of the map features.
Our trained policies also include a small penalty for collisions with environment, equal to $0.01\lambda$.

\section{Evaluation}
%We compare MaAST to several pertinent baselines and demonstrate its effectiveness in PointGoal navigation. 
%We now proceed to describe the evaluation set up, the baselines considered, and the obtained empirical results. 
\noindent\textbf{Simulator and Dataset: }We use Habitat simulator \cite{habitat19iccv}, which supports different datasets such as Matterport3D~\cite{Matterport3D}, Gibson~\cite{Gibsonenv}, \etc. 
We focus our experiments on Matterport3D~\cite{Matterport3D}, which consists of 3D reconstructions of $90$ houses with a total of $2056$ rooms, as it exhibits the greatest episodic complexity compared to the others.
%Semantic segmentation masks are available for all the scenes in this dataset. 
We use standard splits following \cite{habitat19iccv}. It is to be noted that there is no overlap of scenes in the Matterport3D splits.
We refer to \cite{habitat19iccv} for more details.
% Agents used in our experiments have a diameter \SI{0.2}{\meter}, a height of \SI{1.5}{\meter}, and can take four actions: \leftaction, \rightaction, \forwardaction, and \stopaction. 

\vspace{0.05cm}
\noindent\textbf{Evaluation Metrics: }An episode is registered as a success if 
%the agent satisfies the following criteria: 
(i) the agent has taken fewer than $500$ steps, (ii) the distance between the agent and goal is less than \SI{0.2}{\meter}, and (iii) the agent takes \stopaction action to indicate that it has reached goal.
The third condition eliminates a situation when the agent randomly stumbles upon the target and receives %success 
reward. 
%In addition to reporting the agent's success rate over the validation set,
We also report success weighted by path length (\spl) \cite{Anderson2018OnEO}. \spl is a measure of the efficacy of the policy's path taken to the goal compared against the ground truth shortest path.
\iffalse
Let the number of evaluation episodes be $N$, the length of the shortest path $s$, and actual distance covered by the agent $d$. \spl is calculated as follows,
\begin{equation}
\spl = \frac{1}{N} \Sum{\mathbbm{1} \{ i  \text{ is successful}\}  \frac{s_{i}}{\max \left(d_{i}, s_{i}\right)}}{i,1,N}.
\label{eq:SPL}
\end{equation}
\fi
\begin{table}[t]
\vspace{0.15cm}
  \centering
   \caption{Results of our proposed method, compared against several alternatives, evaluated for visual navigation on the Matterport3D validation set. We report both success weighted by path length (\spl) and episode success rate. $\mathbf{\dagger}$: uses our egocentric map Transformer.}
   \vspace{-0.15cm}
  \footnotesize
%   \begin{tabular}{@{} p{0.8in} p{0.6in} cccc@{}}
\scalebox{0.73}{
\begin{tabular}{llccccc}

      \toprule
      & & & \multicolumn{2}{c}{Rewards} & & \\
      \cmidrule{4-5}
      \multicolumn{2}{c}{Method} & Map & Goal & Exploration & \spl(\%) & Succ(\%) \\
      \midrule
    %   \multirow{1}{*}{\rgb}
    %   & RL (PPO) & $\checkmark$ & & & & & & $-$ & $-$ \\
    %   \multirow{1}{*}{\depth}
    %   & RL (PPO) & & $\checkmark$ & & & & & $-$ & $-$ \\
     \rgbd~\cite{habitat19iccv} & RL (PPO) &  & $\checkmark$ & & $34$ & $51$  \\
     \rgbd~\cite{habitat19iccv} & SLAM & &  & & $42$ & - \\
     \rgbdocc~\cite{exp4nav} & RL (PPO) & Occupancy & $\checkmark$ & $\checkmark$ &  $43$ & $52$ \\
      \mast & RL (PPO) & Semantic$\dagger$  & $\checkmark$ & $\checkmark$ & \textbf{47} & \textbf{54} \\
      \bottomrule
  \end{tabular}}

  \label{tab:baselines}
%  \vspace{\captionReduceBot}
\vspace{-0.05cm}
\end{table}

\vspace{0.05cm}
\noindent\textbf{Implementation Details: }Following~\cite{Anderson2018OnEO}, we use the same experimental setup as~\cite{habitat19iccv} for PointGoal navigation. 
%All of the RL-based experiments are trained and evaluated on an Ubuntu machine with 8 GPUs. 
We implemented the networks using PyTorch~\cite{paszke2017automatic} and the \texttt{habitat-sim} and \texttt{habitat-api} modules~\cite{habitat19iccv}. 
For training, we use rollout size of 256, hidden dimension of 512, and learning rate of $10^{-4}$.
For our egocentric map Transformer, we adapt the implementation ~\cite{Wolf2019HuggingFacesTS} as described in \Cref{sec:attention}.
We use a 2-layer BERT model with 4 attention heads each and embeddings of size 128 for both positional and semantic embeddings. 
As in~\cite{habitat19iccv}, we work with a case where our agents are equipped with an idealized GPS sensor and compass. 
Following the use of ground truth scene information (such as ground truth depth) in prior works \cite{exp4nav,habitat19iccv,ddppo,semmapnet}, we utilize the known depth and label of each pixel from the simulation environment to enable a fair comparison.\footnote{However, we experiment below leveraging \mast with online predicted semantics, as well.}

%As in~\cite{habitat19iccv}, we work with a case where our agents are equipped with an idealized GPS sensor and compass to enable fair comparison.

\subsection{Quantitative Results}
%\vspace{-0.1cm}

\subsubsection{Compared Methods}
We compare \mast with several baselines in \Cref{tab:baselines} on the MatterPort3D~\cite{Matterport3D} validation set.

{
\renewcommand{\paragraph}[1]{\smallskip\noindent{\underline{#1}}}
% \underline{Random:}
% The agent takes random actions at each time step $t$ and issues \stopaction when it is within \SI{0.2}{\meter} of the goal.

% \underline{Forward Only:} The agent takes only \forwardaction action and issues \stopaction when it is within \SI{0.2}{\meter} of the goal.

% \underline{Goal Follower:} The agent follows the goal; \ie if the agent's direction is not aligned with the goal, it either performs a \leftaction or \rightaction and issues \stopaction when it is within \SI{0.2}{\meter} of the goal.

\underline{RL (PPO):} The agents are trained end-to-end using DRL, namely PPO, with reward mechanism in \Cref{sec:rewards}. 
We study the performance of RL-based agents with different input sensor modalities:
% We also investigate the additional advantage of our proposed attention mechanism by performing an ablation study with methods not using attention:
\begin{itemize*}[label=\textbf*]
    % \item[\blind,] with no visual sensor input; \ie, the agent only receives information about the goal location and the previous action;
    \item[\rgbd,] with only \rgb and \depth sensors as inputs;
    \item[\rgbdocc,] the approach proposed in \cite{exp4nav} (\ie, along with \rgbd input, the method also uses a egocentric occupancy map), trained with both goal and exploration rewards. In prior work \cite{exp4nav}, the policy architecture was trained exclusively for exploration; however, in our case, it is trained by PPO end-to-end for PointGoal navigation.
    % \item[\rgbdsem,] identical to \rgbdocc, but using the egocentric semantic map with no map Transformer; and
   % \item[\rgbdexp,] like \rgbd but the agent receives additional exploration rewards.
\end{itemize*}
}

% \textcolor{red}{DD-PPO
% %, decentralized distributed PPO algorithm, 
% is a recently proposed DRL algorithm that has been applied to PointGoal navigation~\cite{ddppo}, training agents with more than 2.5 billion steps of experience. 
% However, contrasting our approach with DD-PPO is beyond the scope of this paper as 
% \begin{enumerate*}[label=\roman*)]
%     %\item our objective is to propose an efficient solution under a fixed budget,
%     \item DD-PPO is a DRL algorithm more than a specific visual navigation solution, and
%     \item we can easily incorporate \mast into any RL-based algorithm like DD-PPO to achieve performance boosts from more efficient, large-scale training.
%     %because of \mast's inherent advantage of better exploiting the scene semantics.
% \end{enumerate*}
% Another recent work, Active Neural SLAM (ANS)~\cite{anm20iclr}, has followed the approach of designing a policy for mapping and exploration and then applying the same architecture for PointGoal navigation. 
% However, ANS
% %is both trained in a different setting than ours (limited sensing and agent localization, different simulation domain) and 
% includes several complex modules (a learned mapper, global and local policies) and incorporates additional supervision from the traversable space in the form of a local planner. Our proposed approach (incorporating important semantic scene information into the map) can be incorporated with ~\cite{anm20iclr} as we did with~\cite{exp4nav}. We leave it as a future work.}

While algorithms other than vanilla PPO have been recently applied in the visual navigation space to enable distributed training with large amounts of experience~\cite{ddppo}, a direct comparison with these methods is beyond the scope of this paper as
\begin{enumerate*}[label=\roman*)]
    %\item our objective is to propose an efficient solution under a fixed budget,
    \item DD-PPO is a DRL algorithm more than a specific visual navigation solution, and
    \item we can easily train \mast with any proven RL algorithm to achieve performance boosts from more efficient, large-scale training.
    %because of \mast's inherent advantage of better exploiting the scene semantics.
\end{enumerate*}
Further, while recent works have shown the success of using hierarchical DRL policies paired with analytical planners such as A$^*$ in exploration and navigation~\cite{anm20iclr,occant}, we restrict our focus on learning and improving end-to-end RL policies with simpler modules.
Our proposed approach (incorporating important semantic scene information into the map) can be incorporated with~\cite{anm20iclr} as we did with~\cite{exp4nav}. We leave it as a future work.

% We first directly report \spl and average success rate for baseline methods using no visual sensors; \ie, the Blind agent from \cite{habitat19iccv}. 
% As one would expect, the blind random and forward-only policies perform very poorly as compared to that of the carefully hand-coded goal-following policy. 
% Surprisingly, a blind agent trained by PPO is already able to achieve a modest success rate (35\%).
% All of these policies were trained for 75 million steps with positive rewards when they move closer to goal and negative penalty for collisions. 
% We call these goal-only rewards, and the corresponding reward functions are described in \autoref{sec:rewards}. 
% It can be seen from these experiments that DRL-based methods perform favorably compared to other methods, we now proceed to report only the performance of methods trained via PPO with different visual sensor input modalities and compare against \mast. 

As our main objective is to show \mast's effectiveness under budget constrained situations, we evaluate all the methods under a fixed training budget 
%(with exception of \blind reported directly from~\cite{habitat19iccv}) 
%We consider evaluating the RL agents equipped with different visual sensors after 
of 14 million steps of experience, as this was the point at which a \depth agent has been shown to exceed performance of SLAM~\cite{habitat19iccv}.
 
 %Owing to the observation in~\cite{habitat19iccv} where after training agents using \depth for 14 million steps the agents started outperforming SLAM, we chose to evaluate all methods after training for 14 million steps. 

\subsubsection{Analysis of Results} In \Cref{tab:baselines}, we demonstrate that the proposed approach (\mast) significantly outperforms other methods and baselines, under a much reduced computational load. Agents equipped with no visual sensors (i.e., blind) are expected to show poor performance. Surprisingly, a blind agent~\cite{habitat19iccv} trained by PPO is already able to achieve a modest success (35\%) with 75 million steps of training experience. However, agents equipped with visual sensors show large performance improvement with reduced experience.

Among agents equipped with visual sensors, \mast shows significant improvement over other agents. The improvement is $\mathbf{+13\%}$ absolute ($\mathbf{+38\%}$ relative) in \spl as compared to \rgbd alone and $\mathbf{+4\%}$ absolute ($\mathbf{+9\%}$ relative) as compared to \rgbdocc. In terms of success rate, the performance increase is comparatively less ($+6\%$ relative over \rgbd and $+4\%$ relative over \rgbdocc). This shows that even though most of the methods succeed at nearly the same rate, \mast's policy demonstrates superior efficacy in producing shorter paths to the target, due to our attention mechanism's ability to focus on more relevant parts of the map, such as potential collisions or new areas to explore.
%\footnote{Please refer to the supplementary video for visualizations of map attention.} 
With respect to the statistical significance of our results, we evaluated the performance of \mast averaged over 5 random seeds and obtained \spl of $47.5\pm 0.7$, making the performance gain over other compared methods roughly an order of magnitude greater than the standard deviation of \mast's performance. Furthermore, we note that \mast is able to achieve significant improvements over SLAM in only 14 million training steps, which could not be achieved by \rgbd alone in prior work, even with five times more experience~\cite{habitat19iccv}.

%Next, we compare against a method in which the agent has visual access---in addition to \rgbd inputs---to a top-down, egocentric occupancy map, which we refer to in our experiments as \rgbdocc. In prior work, a similar policy architecture was used to train a model exclusive for exploration of indoor environments~\cite{exp4nav}; however, in our case, it is trained by PPO end-to-end to learn PointGoal navigation.

%While this method results in a small drop in overall success ($-1\%$ as compared to \rgbdexp), it can be seen that visual access to the occupancy map allows for even further gains in path efficiency ($+5\%$ in \spl).

 %In terms of success rate, there is not much of a difference ($+3\%$ over \rgbd), which 

% \paragraph{Computational efficiency:} 

% \paragraph{\rgbd:} 

% As can be seen in \autoref{tab:baselines}, we observe that the proposed approach achieves significant gains \wrt both the evaluation metrics \spl and  average success rate than the multiple baseline state of the art approaches. Our method outperforms the baselines significantly, while our agents were trained for far lesser time and experience (measured in millions of steps).
% Thus our proposed approach results in a $\approx$ 10 fold reduction in  training time while  simultaneously providing considerable performance gains. 

%\vspace{-0.1cm}
\subsection{Ablation Study}
%\vspace{-0.1cm}

To investigate the impact of different components of our proposed MaAST, we perform an ablation study comparing several baselines in \Cref{tab:ablation}. All the methods in \Cref{tab:ablation} are trained with both goal and exploration rewards except \rgbd which is trained with only goal rewards. 
%:\begin{itemize*}[label=\textbf*]
    % \item[\blind,] with no visual sensor input; \ie, the agent only receives information about the goal location and the previous action;
   % \item[\rgbd,] with only \rgb and \depth sensors as inputs and trained with goal rewards; 
   % \item[\rgbdexp,] like \rgbd but the agent receives additional exploration rewards;
    %\item[\rgbdsem,] along with \rgbd input, the method also uses the egocentric semantic map, trained with both goal and exploration rewards;
    %\item[\occatt,] along with \rgbd input, the method uses a egocentric occupancy map and our novel multi-layer egocentric map Transformer, trained with both goal-oriented and exploration rewards.
%\end{itemize*}

\textbf{Exploration Rewards.} We can compare $\rgbd$ and $\rgbdexp$ baselines in \Cref{tab:ablation} to evaluate the impact of the addition of exploration reward.
%$\rgbd$ is trained with \rgb and \depth sensors, using the baseline policy architecture as described in \Cref{sec:policy} and the goal-only rewards.
$\rgbd$ agent performs reasonably well with a $51\%$ success rate. However, as seen in the \spl of \rgbd ($34\%$), the agent takes much longer paths as compared to the ground-truth shortest distance, thus rendering this policy inefficient in terms of path length.
We hypothesize this drop is due to the agent being not able to make efficient use of the implicit semantic cues in the scene.

% \paragraph{\rgbdexp:} 
To encourage the agent to move more efficiently,
%(and to understand better the relative contribution of our proposed methods)
we follow a similar procedure as above to train a policy that utilizes \rgbd along with additional exploration reward: \rgbdexp.
However, the agent has no explicit access to the progressively growing, egocentric occupancy map.
The method achieves a $4\%$ increase in \spl ($38\%$); thus, the addition of exploration reward not only helps the agent reach its goals more often but also takes more efficient paths to do so.

\begin{table}[t]
\vspace{0.2cm}
  \centering
  \footnotesize
   \caption{Ablation study of the proposed \mast on Matterport3D. All models are evaluated after training for 14 million steps of experience using RL (PPO). $\mathbf{\dagger}$: uses our egocentric map Transformer.}
   \vspace{-0.1cm}
%   \begin{tabular}{@{} p{0.8in} p{0.6in} cccc@{}}
\scalebox{0.74}{
\begin{tabular}{lc|c|cc}
      %\toprule \\
      \hline
     \rule{0pt}{1\normalbaselineskip} & Method & Map & \spl(\%) & Succ(\%) \\
      \hline
     \rule{0pt}{1\normalbaselineskip}\multirow{1}{*}{\mast w/o \texttt{(EXP,SEM,ATT)}}
      & \rgbd   &   &  $34$ & $51$ \\
      \multirow{1}{*}{\mast w/o \texttt{(SEM,ATT)}}
      & \rgbdexp &  &  $38$ & $53$ \\
      \multirow{1}{*}{\mast w/o \texttt{ATT}}
      & \rgbdsem & Semantic  & $41$ & $49$ \\
      \multirow{1}{*}{\mast w/o \texttt{SEM}, + \texttt{OCC}}
      & \occatt & Occupancy$\dagger$ & $44$ & $52$ \\
      \multirow{1}{*}{\mast}
      & \texttt{RGBD+SEM+ATT}  & Semantic$\dagger$  & \textbf{47} & \textbf{54} \\
      \hline
  \end{tabular}}
 %}

  \label{tab:ablation}
%  \vspace{\captionReduceBot}
\vspace{-0.1cm}
\end{table}

\textbf{Egocentric Semantic Map.} It is evident from \Cref{tab:ablation} that our use of egocentric semantic map leads to an overall improvement in performance. To analyze the impact, we can compare several variants of \mast (i.e., \occatt vs. \mast, and \rgbdexp vs. \rgbdsem). We observe that \mast achieves ($+3\%$ absolute, $7\%$ relative) improvement in SPL and ($+2\%$ absolute, $4\%$ relative) improvement in success rate using egocentric semantic map compared to \occatt (which uses egocentric occupancy map). However, comparing \rgbdsem to \rgbd, we find that use of semantic map helps \rgbdsem to achieve $+3\%$ improvement in SPL, but there is a drop in performance in terms of success rate ($-2\%$). We suspect this is because the information encoded in the egocentric semantic map is rich, however, the agent is unable to adequately decode and focus on relevant parts of the map (which inspires our use of the Transformer-based map attention mechanism).

\textbf{Real \vs Predicted Semantics.} To better measure the robustness of our method to sensor noise, we conducted an experiment utilizing predicted semantic segmentation, following the map construction method of~\cite{chaplot20objectgoal}; \ie, segment the \rgb observation with Mask R-CNN trained on MSCOCO and project those classes using \depth.
This yields a sparser map now with 80 semantic classes rather than 40 from Matterport3D, and now only filled with confident predictions from the semantic segmentation model rather than dense labels for each pixel.
Due to computational constraints, we train this model for 7.7 million steps of experience, at which point it achieves 34.8\% \spl 
%(\red{NUMBER} success), 
outperforming our \rgbd results with about half as much experience.
This demonstrates the robustness of \mast to even noisy semantic observations and its ability to adapt them to efficient navigation. 

\textbf{Egocentric Map Transformer.}  Our novel attention (\texttt{ATT}) mechanism based on multi-layer egocentric map Transformer leads to significant improvement in performance (\Cref{tab:ablation}). 
%The two best performing models in \Cref{tab:ablation} (\mast and \occatt) uses our proposed map Transformer.
We observe that \mast achieves large performance improvement in SPL, compared to \rgbdexp (+9\%) and \rgbdsem (+6\%), due to the effective use of semantic cues. However, despite having access to the semantic map, \rgbdsem fails to achieve consistent performance improvement over \rgbdexp. This phenomenon supports our hypothesis that egocentric map transformer helps agents attend to most relevant regions of the map for efficient navigation.

We also use the proposed egocentric map Transformer (\Cref{sec:attention}) with the occupancy map (\occatt). However, this
%replacement of $\fmap$ with the map Transformer to perform feature extraction on the egocentric occupancy map 
does not result in a noticeable change in performance. We believe this is because of the information encoded in the egocentric occupancy map is significantly less rich and less diverse compared to the semantic map. Rather than encoding each map cell to about 40 semantic classes in the egocentric semantic map, each occupancy map cell encodes either \emph{unexplored}, \emph{traversable}, and \emph{non-traversable}. 

%Hence, with effective use of egocentric map transformer, \mast is able to achieve significantly more performance improvement compared to \occatt.

%For \occatt, we observe that the replacement of $\fmap$ with the map Transformer to perform feature extraction on the egocentric occupancy map slightly improves \spl with respect to \rgbdocc.

%For \rgbdsem, there is a drop in performance both in terms of success ($-3\%$) and \spl ($-2\%$) as compared to \rgbdocc.
%We suspect this is because the information encoded in the egocentric semantic map is much richer than the occupancy map, so the agent is unable to adequately decode and focus on relevant parts of the map (which inspires our use of the map Transformer).
% or the training budget is insufficient given the baseline network capacity.

%\mast (i.e., \texttt{RGBD+SEM+ATT})

%Finally, \mast, our proposed method, brings together all of the components: \rgb and \depth sensors, a top-down, egocentric semantic map, goal-oriented and exploration rewards, and our novel multi-layer egocentric map Transformer on the semantic map input.
% \paragraph{MaAST:} 

\vspace{-0.1cm}
\subsection{Qualitative Results}

\begin{figure}
\vspace{0.2cm}
    \newlength{\mapwidth}
    \setlength{\mapwidth}{0.48\textwidth}
    \centering
    \includegraphics[width=\mapwidth, height=0.145\textwidth]{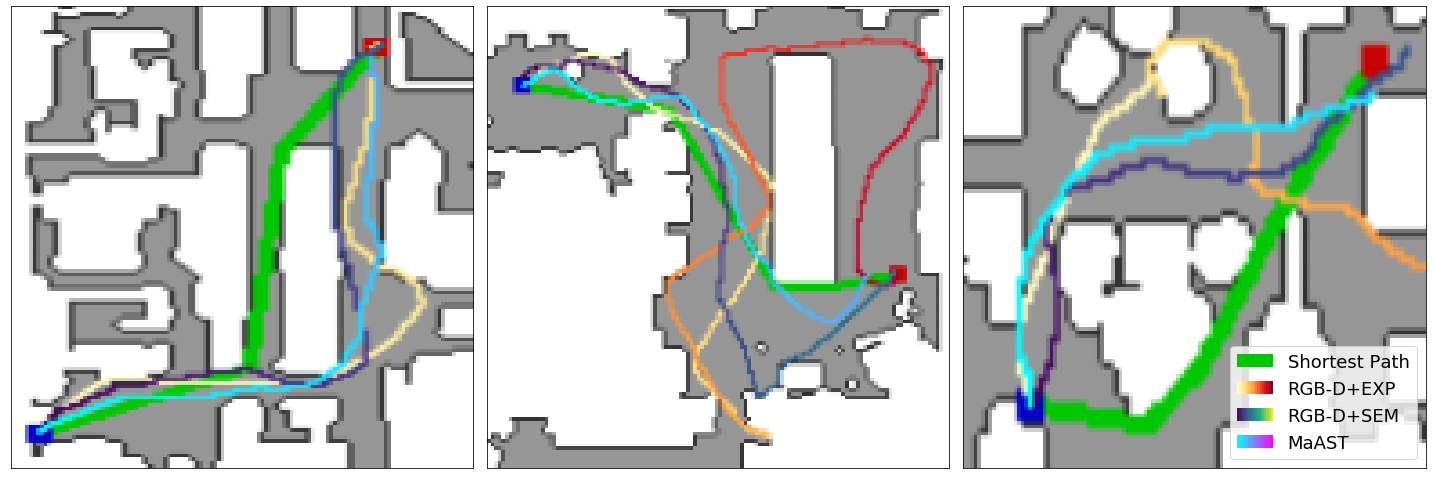}
    \begin{minipage}{0.3\mapwidth}
        \subcaption{}\label{qual11}
    \end{minipage}
    \begin{minipage}{0.3\mapwidth}
        \subcaption{}\label{qual12}
    \end{minipage}
    \begin{minipage}{0.3\mapwidth}
        \subcaption{}\label{qual13}
    \end{minipage}\\
    \vspace{-0.1cm}
    \includegraphics[width=\mapwidth, height=0.145\textwidth]{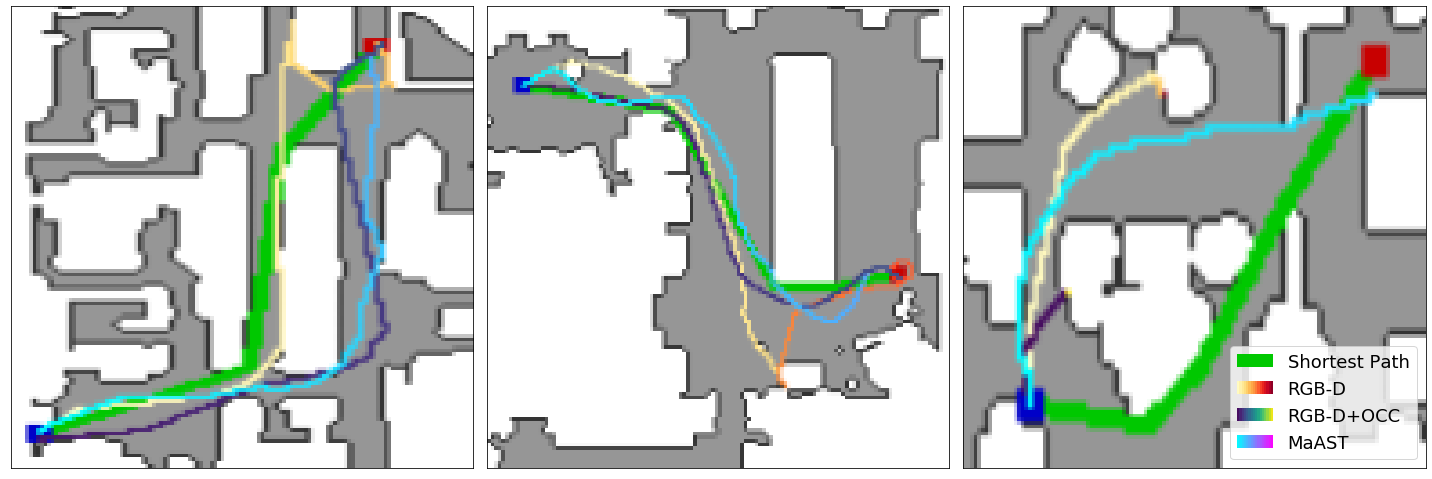}
    \begin{minipage}{0.3\mapwidth}
        \subcaption{}\label{qual21}
    \end{minipage}
    \begin{minipage}{0.3\mapwidth}
        \subcaption{}\label{qual22}
    \end{minipage}
    \begin{minipage}{0.3\mapwidth}
        \subcaption{}\label{qual23}
    \end{minipage}
    \vspace{-0.3cm}
    \caption{A visualization of the path taken by different methods for three episodes from Matterport3D. The ground truth shortest path is shown by a solid green line. The agents' start and goal locations are marked with a blue and red dot, respectively. The color of the agents' paths shifts along a gradient (reproduced in the legend) as the maximum number of steps for the episode (500) is approached.}
    \label{fig:examples}
    \vspace{-0.1cm}
\end{figure}

In \Cref{fig:examples}, we provide some visualizations of the paths toward the goal followed by the various methods in three different episodes from the Matterport3D validation set.
In the top row (\Cref{qual11,qual12,qual13}), we compare \mast with \rgbdexp, 
%trained with exploration rewards but without a map, 
and \rgbdsem.
%, trained with the map but without the proposed attention mechanism.
In the bottom row (\Cref{qual21,qual22,qual23}), we compare against our two baselines from prior work, \rgbd and \rgbdocc.
%The shortest path (by geodesic distance) is shown with a bold green line.
All methods have their path colored with a gradient to capture the number of steps taken by the agent, all scaled between 0 and 500 steps.
For example, in \Cref{qual22}, \rgbd agent moves quickly towards the goal at first; however, it becomes stuck behind a chair.
When it moves away, its path is much more darkly colored, to indicate the amount of time taken to recover from the obstacle.

In the first episode (\Cref{qual11,qual21}), 
%most of the agents perform quite well, all reaching 
all the agents reach the goal, but the \rgbd and \rgbdexp take very long paths to do so.
It is clear 
%from these examples 
that \rgbdexp tends to very aggressively follow the available traversable space, which leads to inefficient paths (\Cref{qual12}) or missing the goal entirely (\Cref{qual13}).
\Cref{qual12,qual13,qual21,qual22} all help to demonstrate the efficiency gains of the \mast attention mechanism.
Even in cases where several of the methods follow similar paths to the goal, \mast is more quickly able to find obscured free space (such as around the obstacle in the center of the room in \Cref{qual12}) to move more efficiently towards the goal.

The final episode (\Cref{qual13,qual23}) represents a failure case for all the methods.
\mast and \rgbdsem reach close to the goal; however the episodes are registered as failures, as neither are within the \SI{0.2}{\meter} success radius.
On the other hand, \rgbdexp completely misses the goal, heading off the visualized map into the next room, while both \rgbd and \rgbdocc get stuck on the furniture and terminate.
In our analysis, we find that for this particular Matterport3D house model, there appears to be holes in the 3D model towards and around the goal, which provide confusing features to the policy network; \eg, all of the furniture at the bottom is fused together.
However, one of the strengths of leveraging semantics in our models seems to be avoiding areas that either present narrow spaces with high collision potential or low-lying objects below the agent's field of view.
This is a instance when we observe the agent utilizing semantics of the environment (\ie, ``furniture is difficult to navigate through'') beyond just which areas are free and which are not. 

\section{Conclusion}
In this paper, we introduced principled techniques for improved performance on visual navigation tasks with autonomous agents, by leveraging rich semantic features while simultaneously operating under a computational budget.
Our proposed \mast, incorporates a novel attention mechanism based on multi-layer Transformers, which encourages agents to focus on the most relevant areas of the egocentric semantic map for navigation. 
Through systematic experimentation with quantitative and qualitative analysis, we demonstrate the superior performance of our proposed approach as compared with several baselines. We show that \mast provides significant performance gains (a $13\%$ gain in path efficiency over purely visual methods), while also outperforming classical geometry-based methods with decreased training budget (time, computational requirements, and cost).

%\textcolor{red}{Future work includes extending \mast to real-world environments where ground truth scene information is not available. We also plan to evaluate the performance of our approach on other navigation tasks, such as ObjectGoal navigation.}
%%%%%%%%%%%%%%%%%%%%%%%%%%%%%%%%%%%%%%%%%%%%%%%%%%%%%%%%%%%%%%%%%%%%%%%%%%%%%%%%

%%%%%%%%%%%%%%%%%%%%%%%%%%%%%%%%%%%%%%%%%%%%%%%%%%%%%%%%%%%%%%%%%%%%%%%%%%%%%%%%

{\small
\bibliographystyle{IEEEtran.bst}
\bibliography{egbib}

\begin{thebibliography}{10}
\providecommand{\url}[1]{#1}
\csname url@rmstyle\endcsname
\providecommand{\newblock}{\relax}
\providecommand{\bibinfo}[2]{#2}
\providecommand\BIBentrySTDinterwordspacing{\spaceskip=0pt\relax}
\providecommand\BIBentryALTinterwordstretchfactor{4}
\providecommand\BIBentryALTinterwordspacing{\spaceskip=\fontdimen2\font plus
\BIBentryALTinterwordstretchfactor\fontdimen3\font minus
  \fontdimen4\font\relax}
\providecommand\BIBforeignlanguage[2]{{%
\expandafter\ifx\csname l@#1\endcsname\relax
\typeout{** WARNING: IEEEtran.bst: No hyphenation pattern has been}%
\typeout{** loaded for the language `#1'. Using the pattern for}%
\typeout{** the default language instead.}%
\else
\language=\csname l@#1\endcsname
\fi
#2}}

\bibitem{braincognition}
N.~Burgess, ``Spatial cognition and the brain,'' \emph{Annals of the New York
  Academy of Sciences}, vol. 1124, no.~1, pp. 77--97, 2008.

\bibitem{humannavigation}
M.~Denis and J.~M. Loomis, ``Perspectives on human spatial cognition: memory,
  navigation, and environmental learning,'' \emph{Psychological Research},
  vol.~71, no.~3, pp. 235--239, 2007.

\bibitem{thrun2005probabilistic}
S.~Thrun, W.~Burgard, and D.~Fox, \emph{Probabilistic robotics}.\hskip 1em plus
  0.5em minus 0.4em\relax MIT press, 2005.

\bibitem{davison1998mobile}
A.~J. Davison and D.~W. Murray, ``Mobile robot localisation using active
  vision,'' in \emph{ECCV}, 1998, pp. 809--825.

\bibitem{desouza2002vision}
G.~N. DeSouza and A.~C. Kak, ``Vision for mobile robot navigation: A survey,''
  \emph{IEEE transactions on pattern analysis and machine intelligence},
  vol.~24, no.~2, pp. 237--267, 2002.

\bibitem{lavalle2000rapidly}
S.~M. LaValle and J.~J. Kuffner~Jr, ``Rapidly-exploring random trees: Progress
  and prospects,'' 2000.

\bibitem{canny1988complexity}
J.~Canny, \emph{The complexity of robot motion planning}.\hskip 1em plus 0.5em
  minus 0.4em\relax MIT press, 1988.

\bibitem{kavraki1996probabilistic}
L.~E. Kavraki, P.~Svestka, J.-C. Latombe, and M.~H. Overmars, ``Probabilistic
  roadmaps for path planning in high-dimensional configuration spaces,''
  \emph{IEEE transactions on Robotics and Automation}, vol.~12, no.~4, pp.
  566--580, 1996.

\bibitem{habitat19iccv}
M.~Savva, A.~Kadian, O.~Maksymets, Y.~Zhao, E.~Wijmans, B.~Jain, J.~Straub,
  J.~Liu, V.~Koltun, J.~Malik, D.~Parikh, and D.~Batra, ``Habitat: {A}
  {P}latform for {E}mbodied {AI} {R}esearch,'' in \emph{ICCV}, 2019.

\bibitem{ddppo}
E.~Wijmans, A.~Kadian, A.~Morcos, S.~Lee, I.~Essa, D.~Parikh, M.~Savva, and
  D.~Batra, ``Decentralized distributed ppo: Solving pointgoal navigation,''
  \emph{arXiv preprint arXiv:1911.00357}, 2019.

\bibitem{vaswani17:attention}
A.~Vaswani, N.~Shazeer, N.~Parmar, J.~Uszkoreit, L.~Jones, A.~N. Gomez,
  L.~Kaiser, and I.~Polosukhin, ``Attention is all you need,'' in \emph{NIPS},
  2017, pp. 5998--6008.

\bibitem{exp4nav}
T.~Chen, S.~Gupta, and A.~Gupta, ``Learning exploration policies for
  navigation,'' in \emph{ICLR}, 2019.

\bibitem{jaderberg2016reinforcement}
M.~Jaderberg, V.~Mnih, W.~M. Czarnecki, T.~Schaul, J.~Z. Leibo, D.~Silver, and
  K.~Kavukcuoglu, ``Reinforcement learning with unsupervised auxiliary tasks,''
  \emph{arXiv preprint arXiv:1611.05397}, 2016.

\bibitem{pathak2017curiosity}
D.~Pathak, P.~Agrawal, A.~A. Efros, and T.~Darrell, ``Curiosity-driven
  exploration by self-supervised prediction,'' in \emph{CVPR workshops}, 2017,
  pp. 16--17.

\bibitem{gupta2017unifying}
S.~Gupta, D.~Fouhey, S.~Levine, and J.~Malik, ``Unifying map and landmark based
  representations for visual navigation,'' \emph{arXiv preprint
  arXiv:1712.08125}, 2017.

\bibitem{mnih2016asynchronous}
V.~Mnih, A.~P. Badia, M.~Mirza, A.~Graves, T.~Lillicrap, T.~Harley, D.~Silver,
  and K.~Kavukcuoglu, ``Asynchronous methods for deep reinforcement learning,''
  in \emph{ICML}, 2016, pp. 1928--1937.

\bibitem{AI2THOR}
E.~Kolve, R.~Mottaghi, D.~Gordon, Y.~Zhu, A.~Gupta, and A.~Farhadi, ``Ai2-thor:
  An interactive 3d environment for visual ai,'' \emph{arXiv preprint
  arXiv:1712.05474}, 2017.

\bibitem{dosovitskiy2016learning}
A.~Dosovitskiy and V.~Koltun, ``Learning to act by predicting the future,''
  \emph{arXiv preprint arXiv:1611.01779}, 2016.

\bibitem{anm20iclr}
\BIBentryALTinterwordspacing
D.~S. Chaplot, D.~Gandhi, S.~Gupta, A.~Gupta, and R.~Salakhutdinov, ``Learning
  to explore using active neural {SLAM},'' in \emph{8th International
  Conference on Learning Representations, {ICLR} 2020, Addis Ababa, Ethiopia,
  April 26-30, 2020}.\hskip 1em plus 0.5em minus 0.4em\relax OpenReview.net,
  2020. [Online]. Available: \url{https://openreview.net/forum?id=HklXn1BKDH}
\BIBentrySTDinterwordspacing

\bibitem{MINOS}
M.~Savva, A.~X. Chang, A.~Dosovitskiy, T.~A. Funkhouser, and V.~Koltun,
  ``Minos: Multimodal indoor simulator for navigation in complex
  environments,'' \emph{arXiv preprint arXiv:1712.03931}, 2017.

\bibitem{brockman2016openai}
G.~Brockman, V.~Cheung, L.~Pettersson, J.~Schneider, J.~Schulman, J.~Tang, and
  W.~Zaremba, ``{OpenAI} gym,'' \emph{arXiv preprint arXiv:1606.01540}, 2016.

\bibitem{Gibsonenv}
F.~Xia, A.~R. Zamir, Z.-Y. He, A.~Sax, J.~Malik, and S.~Savarese, ``{Gibson
  Env}: Real-world perception for embodied agents,'' in \emph{CVPR}, 2018, pp.
  9068--9079.

\bibitem{vizdoom}
M.~Kempka, M.~Wydmuch, G.~Runc, J.~Toczek, and W.~Ja{\'s}kowski, ``Vizdoom: A
  doom-based ai research platform for visual reinforcement learning,'' in
  \emph{2016 IEEE Conference on Computational Intelligence and Games
  (CIG)}.\hskip 1em plus 0.5em minus 0.4em\relax IEEE, 2016, pp. 1--8.

\bibitem{chalet}
C.~Yan, D.~Misra, A.~Bennnett, A.~Walsman, Y.~Bisk, and Y.~Artzi, ``Chalet:
  Cornell house agent learning environment,'' \emph{arXiv preprint
  arXiv:1801.07357}, 2018.

\bibitem{SUNCG}
S.~Song, F.~Yu, A.~Zeng, A.~X. Chang, M.~Savva, and T.~Funkhouser, ``Semantic
  scene completion from a single depth image,'' in \emph{CVPR}, 2017, pp.
  1746--1754.

\bibitem{stanfordscenes}
M.~Fisher, D.~Ritchie, M.~Savva, T.~Funkhouser, and P.~Hanrahan,
  ``Example-based synthesis of 3d object arrangements,'' \emph{ACM Transactions
  on Graphics (TOG)}, vol.~31, no.~6, p. 135, 2012.

\bibitem{scenenet}
A.~Handa, V.~Patraucean, V.~Badrinarayanan, S.~Stent, and R.~Cipolla,
  ``Understanding real world indoor scenes with synthetic data,'' in
  \emph{CVPR}, 2016, pp. 4077--4085.

\bibitem{Matterport3D}
A.~X. Chang, A.~Dai, T.~A. Funkhouser, M.~Halber, M.~Nie{\ss}ner, M.~Savva,
  S.~Song, A.~Zeng, and Y.~Zhang, ``Matterport3d: Learning from rgb-d data in
  indoor environments,'' \emph{2017 International Conference on 3D Vision
  (3DV)}, pp. 667--676, 2017.

\bibitem{stanford2d3d}
I.~Armeni, O.~Sener, A.~R. Zamir, H.~Jiang, I.~Brilakis, M.~Fischer, and
  S.~Savarese, ``3d semantic parsing of large-scale indoor spaces,'' in
  \emph{CVPR}, 2016, pp. 1534--1543.

\bibitem{Mishkin2019BenchmarkingCA}
D.~Mishkin, A.~Dosovitskiy, and V.~Koltun, ``Benchmarking classic and learned
  navigation in complex 3d environments,'' \emph{arXiv preprint
  arXiv:1901.10915}, 2019.

\bibitem{MidLevelVisualRep}
A.~Sax, B.~Emi, A.~R. Zamir, L.~J. Guibas, S.~Savarese, and J.~Malik,
  ``Mid-level visual representations improve generalization and sample
  efficiency for learning active tasks,'' \emph{arXiv preprint
  arXiv:1812.11971}, 2018.

\bibitem{situationalfusion}
W.~B. Shen, D.~Xu, Y.~Zhu, L.~J. Guibas, F.~Li, and S.~Savarese, ``Situational
  fusion of visual representation for visual navigation,'' in \emph{ICCV},
  2019.

\bibitem{yamauchi1997frontier}
B.~Yamauchi, ``A frontier-based approach for autonomous exploration.'' in
  \emph{cira}, vol.~97, 1997, p. 146.

\bibitem{thrun1999minerva}
S.~Thrun, M.~Bennewitz, W.~Burgard, A.~B. Cremers, F.~Dellaert, D.~Fox,
  D.~Hahnel, C.~Rosenberg, N.~Roy, J.~Schulte, \emph{et~al.}, ``Minerva: A
  second-generation museum tour-guide robot,'' in \emph{ICRA}, vol.~3.\hskip
  1em plus 0.5em minus 0.4em\relax IEEE, 1999.

\bibitem{savinov2018episodic}
N.~Savinov, A.~Raichuk, R.~Marinier, D.~Vincent, M.~Pollefeys, T.~Lillicrap,
  and S.~Gelly, ``Episodic curiosity through reachability,'' \emph{arXiv
  preprint arXiv:1810.02274}, 2018.

\bibitem{gupta17:cogmapping}
S.~Gupta, J.~Davidson, S.~Levine, R.~Sukthankar, and J.~Malik, ``Cognitive
  mapping and planning for visual navigation,'' in \emph{CVPR}, 2017, pp.
  7272--7281.

\bibitem{occant}
\BIBentryALTinterwordspacing
S.~K. Ramakrishnan, Z.~Al{-}Halah, and K.~Grauman, ``Occupancy anticipation for
  efficient exploration and navigation,'' in \emph{Computer Vision - {ECCV}
  2020 - 16th European Conference, Glasgow, UK, August 23-28, 2020,
  Proceedings, Part {V}}, ser. Lecture Notes in Computer Science, A.~Vedaldi,
  H.~Bischof, T.~Brox, and J.~Frahm, Eds., vol. 12350.\hskip 1em plus 0.5em
  minus 0.4em\relax Springer, 2020, pp. 400--418. [Online]. Available:
  \url{https://doi.org/10.1007/978-3-030-58558-7\_24}
\BIBentrySTDinterwordspacing

\bibitem{chaplot20objectgoal}
\BIBentryALTinterwordspacing
D.~S. Chaplot, D.~Gandhi, A.~Gupta, and R.~Salakhutdinov, ``Object goal
  navigation using goal-oriented semantic exploration,'' \emph{CoRR}, vol.
  abs/2007.00643, 2020. [Online]. Available:
  \url{https://arxiv.org/abs/2007.00643}
\BIBentrySTDinterwordspacing

\bibitem{Yang2018VisualSN}
W.~Yang, X.~Wang, A.~Farhadi, A.~Gupta, and R.~Mottaghi, ``Visual semantic
  navigation using scene priors,'' \emph{arXiv preprint arXiv:1810.06543},
  2018.

\bibitem{NeuralMap}
E.~Parisotto and R.~Salakhutdinov, ``Neural map: Structured memory for deep
  reinforcement learning,'' \emph{arXiv preprint arXiv:1702.08360}, 2017.

\bibitem{Oh2016ControlOM}
J.~Oh, V.~Chockalingam, S.~P. Singh, and H.~Lee, ``Control of memory, active
  perception, and action in minecraft,'' \emph{arXiv preprint
  arXiv:1605.09128}, 2016.

\bibitem{Anderson2018OnEO}
P.~Anderson, A.~X. Chang, D.~S. Chaplot, A.~Dosovitskiy, S.~Gupta, V.~Koltun,
  J.~Kosecka, J.~Malik, R.~Mottaghi, M.~Savva, and A.~R. Zamir, ``On evaluation
  of embodied navigation agents,'' \emph{arXiv preprint arXiv:1807.06757},
  2018.

\bibitem{turbo}
A.~Mikhailov, ``Turbo, an improved rainbow colormap for visualization,''
  \url{https://ai.googleblog.com/2019/08/turbo-improved-rainbow-colormap-for.html},
  2019.

\bibitem{Resnet}
K.~He, X.~Zhang, S.~Ren, and J.~Sun, ``Deep residual learning for image
  recognition,'' in \emph{CVPR}, 2016, pp. 770--778.

\bibitem{BERT}
J.~Devlin, M.-W. Chang, K.~Lee, and K.~Toutanova, ``Bert: Pre-training of deep
  bidirectional transformers for language understanding,'' in \emph{NAACL-HLT},
  2019.

\bibitem{selfatt}
H.~Zhang, I.~J. Goodfellow, D.~N. Metaxas, and A.~Odena, ``Self-attention
  generative adversarial networks,'' in \emph{{ICML}}, ser. Proceedings of
  Machine Learning Research, vol.~97.\hskip 1em plus 0.5em minus 0.4em\relax
  {PMLR}, 2019, pp. 7354--7363.

\bibitem{visualbert}
L.~H. Li, M.~Yatskar, D.~Yin, C.~Hsieh, and K.~Chang, ``Visualbert: {A} simple
  and performant baseline for vision and language,'' \emph{arXiv preprint
  arXiv:1908.03557}, 2019.

\bibitem{vilbert}
J.~Lu, D.~Batra, D.~Parikh, and S.~Lee, ``Vilbert: Pretraining task-agnostic
  visiolinguistic representations for vision-and-language tasks,'' \emph{arXiv
  preprint arXiv:1908.02265}, 2019.

\bibitem{cho14:gru}
K.~Cho, B.~van Merrienboer, {\c{C}}.~G{\"{u}}l{\c{c}}ehre, D.~Bahdanau,
  F.~Bougares, H.~Schwenk, and Y.~Bengio, ``Learning phrase representations
  using {RNN} encoder-decoder for statistical machine translation,'' in
  \emph{EMNLP}, 2014, pp. 1724--1734.

\bibitem{ppo}
J.~Schulman, F.~Wolski, P.~Dhariwal, A.~Radford, and O.~Klimov, ``Proximal
  policy optimization algorithms,'' \emph{arXiv preprint arXiv:1707.06347},
  2017.

\bibitem{paszke2017automatic}
A.~Paszke, S.~Gross, S.~Chintala, G.~Chanan, E.~Yang, Z.~DeVito, Z.~Lin,
  A.~Desmaison, L.~Antiga, and A.~Lerer, ``Automatic differentiation in
  {PyTorch},'' in \emph{NIPS Autodiff Workshop}, 2017.

\bibitem{Wolf2019HuggingFacesTS}
T.~Wolf, L.~Debut, V.~Sanh, J.~Chaumond, C.~Delangue, A.~Moi, P.~Cistac,
  T.~Rault, R.~Louf, M.~Funtowicz, and J.~Brew, ``Huggingface's transformers:
  State-of-the-art natural language processing,'' \emph{ArXiv}, vol.
  abs/1910.03771, 2019.

\bibitem{semmapnet}
\BIBentryALTinterwordspacing
V.~Cartillier, Z.~Ren, N.~Jain, S.~Lee, I.~Essa, and D.~Batra, ``Semantic
  mapnet: Building allocentric semanticmaps and representations from egocentric
  views,'' \emph{CoRR}, vol. abs/2010.01191, 2020. [Online]. Available:
  \url{https://arxiv.org/abs/2010.01191}
\BIBentrySTDinterwordspacing

\end{thebibliography}
}

\end{document}